%% file: acl_latex.tex
\pdfoutput=1
\documentclass[11pt]{article}
\usepackage{acl}

\usepackage{times}
\usepackage{latexsym}
\usepackage{listings}
\usepackage{amsthm}
\usepackage{tabularx}
\usepackage{graphicx}
\usepackage[T1]{fontenc}

\usepackage[utf8]{inputenc}

\usepackage{microtype}

\usepackage{inconsolata}

\title{Aspect-Based Sentiment Analysis for Open-Ended HR Survey Responses}

\author{Lois Rink\thanks{Work done while on internship at Randstad Groep Nederland} \\
  Universiteit van Amsterdam \\
  Amsterdam, The Netherlands \\
  \texttt{Lrink@hotmail.com} \\\And
  Job Meijdam \\
  Randstad Groep Nederland \\
  Diemen, The Netherlands \\
  \texttt{job.meijdam@randstadgroep.nl} \\\AND
  David Graus \\
  Randstad \\
  Diemen, The Netherlands \\
  \texttt{david.graus@randstad.com} \\}

\begin{document}
\maketitle
\begin{abstract}
Understanding preferences, opinions, and sentiment of the workforce is paramount for effective employee lifecycle management. 
Open-ended survey responses serve as a valuable source of information. 
This paper proposes a machine learning approach for aspect-based sentiment analysis (ABSA) of Dutch open-ended responses in employee satisfaction surveys. 
Our approach aims to overcome the inherent noise and variability in these responses, enabling a comprehensive analysis of sentiments that can support employee lifecycle management. 
Through response clustering we identify six key aspects (salary, schedule, contact, communication, personal attention, agreements), which we validate by domain experts. 
We compile a dataset of 1,458 Dutch survey responses, revealing label imbalance in aspects and sentiments. 
We propose few-shot approaches for ABSA based on Dutch BERT models, and compare them against bag-of-words and zero-shot baselines.
Our work significantly contributes to the field of ABSA by demonstrating the first successful application of Dutch pre-trained language models to aspect-based sentiment analysis in the domain of human resources (HR).
\end{abstract}

\input{sections/1-introduction.tex}
\input{sections/2-related-work.tex}
\input{sections/3-methodology.tex}
\input{sections/4-results.tex}
\input{sections/5-discussion.tex}
\input{sections/6-conclusion.tex}

\section*{Acknowledgements}
The authors would like to thank PP for extensive infrastructural support in running our annotation study. 

\bibliography{custom}

\end{document}

%% file: sections/1-introduction.tex
\section{Introduction}
\label{sec:introduction}
Understanding employees' preferences and opinions can be of paramount importance in the full employee life cycle, e.g., from recruitment and selection to employee retention, and performance and career management~\cite{10.1145/3523227.3547414}. 
In employee satisfaction surveys, open-ended questions may elicit a wide range of aspects. 
However, conducting large-scale analysis of these responses is challenging because of their user-generated nature. 

Aspect-based sentiment analysis (ABSA) is the task of identifying and extracting sentiments toward specific aspects from free text, allowing a more detailed analysis of opinions~\cite{pontiki2016semeval}, which means they can be a valuable tool for identifying specific areas of (dis)satisfaction. 

Despite the promise of ABSA, limited research has explored its application beyond English~\cite{nazir2020issues}. 
Moreover, to our knowledge, ABSA has not been studied in employee satisfaction surveys. 
This study aims to bridge this research gap by studying ABSA on Dutch open-ended employee satisfaction survey responses. 
The central research question that we answer in this paper is:
\begin{itemize}
    \item[RQ1] \textit{How effective is a BERT-based machine learning model in extracting aspect-sentiments from Dutch open-ended responses of employee satisfaction surveys?}
\end{itemize}

To answer this question, we first aim to answer the following sub-questions:
\begin{itemize}
    \item[RQ1.1] \textit{How does the performance of few-shot classification using Dutch BERT models compare to bag-of-words based baselines in ABSA?}
    \item[RQ1.2] \textit{How does the performance of few-shot classification using Dutch BERT models compare to a zero-shot classification baseline in ABSA?}
    \item[RQ1.3] \textit{To what extent can an improvement in performance be achieved in few-shot aspect-based sentiment classification, by training on a data set enlarged through data augmentation?}
\end{itemize}

Prior work mainly focuses on ABSA in English microblogs and user reviews. 
Microblogs, like X, cover diverse topics using informal language with an assumed shared context. Reviews tend to adopt more formal styles and primarily revolve around specific products or services~\cite{Kumar2019Sentiment}.

Open-ended survey responses share similarities with microblogs in terms of writing and context, but have topic coverage narrowed by topics that affect employee (dis)satisfaction. 

This study compares the performance of Dutch pre-trained language models, BERTje \cite{devries2019bertje} and RobBERT \cite{delobelle2020robbert}, for ABSA in a few-shot classification experiment. These BERT-based models leverage contextual information, which is advantageous for short texts with limited contextual cues \cite{chang2020taming}. We assess their performance against zero-shot classification BERTje and RobBERT models and traditional bag-of-words models~\cite{Wu2020Identifying}.

The specific contributions of our research are as follows:
\begin{enumerate}
    \item Annotation of a data set of 1,458 open-ended survey responses for ABSA in Randstad, with publicly available annotation procedures and guidelines for future studies.
    \item Development of an ABSA model for Dutch open-ended employee satisfaction survey responses, enabling automated aspect-sentiment extraction for efficient and accurate analysis.
\end{enumerate}

%% file: sections/2-related-work.tex
\section{Related Work}
\label{sec:related_work}
This section provides an overview of previous studies on aspect and  sentiment classification, ABSA, and the Dutch BERT models we employ in this paper: BERTje and RobBERT.

\subsection{Aspect classification}
Recent studies have found transformer models' effectiveness in topic and aspect classification on short text through their ability to capture long-range dependencies and context.
\citet{chang2020taming} demonstrated fine-tuning a deep transformer network for extreme multi-label aspect classification in English. 
In contrast, \citet{dadgar2016novel} utilized a combination of TF-IDF vectors and an SVM classifier for news article aspect detection, without requiring extensive training. 
\citet{hu2021multi} demonstrated few-shot learning using prototypical networks for aspect classification is valuable when labelled data is scarce. 
Alternatively, zero-shot classification, as discussed by \citet{yin2019benchmarking}, allows topic classification without specific training, relying solely on labelled data for validation.

\subsection{Sentiment classification}
Sentiment classification is a long-standing research focus. 
\citet{jimenez2017analysis} used an SVM classifier for patient satisfaction categorization in Dutch and Spanish healthcare reviews. 
\citet{karl2022transformers} demonstrated that larger transformer models, like RoBERTa, excel in sentiment classification over classic BERT models due to their ability to generalize to unseen data. 
These transformer models can also perform well in zero-shot multilingual sentiment classification, as shown by \citet{tesfagergish2022zero}.
\citet{dogra2021analyzing} illustrated the effectiveness of BERT-based models in few-shot sentiment classification. 
Between Dutch BERT models it was found that RobBERT outperforms BERTje in sentiment classification, credited to its enhanced training framework and a larger training corpus~\cite{de2021emotional}.

\subsection{Aspect-based sentiment analysis}
\citet{lin2009joint} introduced joint aspect-sentiment analysis, combining LDA for aspect extraction and a polarity lexicon for sentiment classification in English movie reviews. However, \citet{jimenez2017analysis} found this lexicon-based approach unsuitable for Dutch and Spanish.

ABSA gained prominence after SemEval-2014 task 4~\cite{pontiki2014semeval}, where researchers tackled identifying explicit terms or categories representing aspects of a target entity, and their polarities, in the context of restaurant and laptop reviews. 
Over time, the task expanded to encompass full and multilingual reviews. 

\citet{de2016rude} attempted Dutch ABSA using SVMs, augmenting their bag-of-words (BoW) approach with semantic role labels, with limited success. 

Recent advancements include \citet{hoang2019aspect} demonstrating the potential of BERT models in English ABSA, and \citet{liao2021improved} improving performance with a RoBERTa-based model. 
Few-shot BERT classification and augmented training with BERT embeddings were explored by \citet{hosseini2022generative}. \citet{de2022sentemo} employed a RobBERT model to extract features for SVM in ABSA, reporting better results than a full transformer-based approach for their pipeline.

ABSA in open-text survey responses was addressed by \citet{cammel2020automatically} and \citet{van2022analyzing} using techniques like LDA, rule-based methods, and non-negative matrix factorization with multilingual BERT. 
These studies primarily focused on patient survey questions in the healthcare domain. 
Additionally, \citet{cammel2020automatically} restricted to detecting a single aspect-sentiment per response, limiting broad insights. 
Both studies considered open-ended responses to different survey questions simultaneously, some of which elicited one-word responses which provided insufficient context for successful aspect-sentiment extraction.

\subsection{Transformer models for Dutch}
Transformers have excelled in Dutch aspect and sentiment classification, with two dedicated to Dutch: BERTje and RobBERT. 

BERTje, with 12 layers, a hidden size of 768, and 12 attention heads, is pre-trained on a diverse corpus encompassing Wikipedia, news articles, books, and web pages, enabling it to capture Dutch linguistic patterns and context~\cite{devries2019bertje}.
RobBERT, a Dutch variant of RoBERTa, shares a similar architecture with BERTje, but benefits from a more extensive pre-training dataset, including Wikipedia, news articles, web pages, Dutch parliament debates, and social media~\cite{delobelle2020robbert}. 

Existing methods for Dutch sentiment analysis are suboptimal, requiring further exploration, especially in novel domains like HR surveys.

%% file: sections/3-methodology.tex
\section{Methodology}
\label{sec:methodology}
This section outlines the development and evaluation of the proposed ABSA model for Dutch open-ended survey responses on employee satisfaction. 

\subsection{Data set}
The data set used in this study comprises Dutch open-ended responses to survey questions conducted by \emph{anonymized}. 
We derived a sample of $1,500$ responses, recorded between January 2019 and December 2022, through stratified sampling across three different sub-brands of Randstad.

An example of such a response which illustrates the challenging nature of ABSA in employee satisfaction survey responses is: \textit{"Ik ben tevreden over mijn salaris. Ik mis wel een stukje persoonlijke aandacht."} ("I am satisfied with my salary. However, I do miss some personal attention.")
Here, we distinguish two aspect-sentiment pairs; an employee expresses a positive sentiment toward their salary, yet a negative sentiment toward personal attention.
We explain the range of identified aspects in Section~\ref{sec:aspect_selection}.

\subsection{Data Preparation}
We filtered and anonymized our set of responses to ensure data quality and privacy. 
We excluded responses with less than 10 tokens, as manual inspection revealed how shorter responses often lacked adequate contextual information for accurate aspect classification. 
We also excluded responses exceeding 512 characters to address computational constraints~\cite{liao2021improved}.
The final average response length is 35.7 tokens (approximately 182 characters), with 10 tokens (39 characters) at a minimum, and 97 tokens (511 characters) at most. 

To ensure anonymity of both respondents and individuals mentioned in responses, all personal information was removed and replaced by dummy variables by using the Dutch Named Entity Recognition (NER) SpaCy model and regular expressions.
After identifying, we replaced person names with “Naam”, email addresses with “Emailadres”, and addresses with “Adres”. 
After this, manual review corrected an additional 28 names missed by SpaCy. 

\subsection{Aspect selection}\label{sec:aspect_selection}
We preprocessed responses by retaining only nouns, proper nouns, and verbs because they convey the most informative content~\cite{boguraev1999dynamic}. 
We then applied lemmatization. 
Finally, we applied TF-IDF vectorization to represent each response.
We then applied $k$-means clustering over these TF-IDF vectors, determining the optimal number of clusters at $k=6$, using the elbow method. 

We inspected the responses in each cluster.
Table \ref{tab:toptermspercluster}  shows the most important terms per cluster, indicated by the highest TF-IDF score. 
After analyzing these clusters, we found they roughly represent the aspects of 
\textit{contact, salary, schedule, personal attention, communication,} and \textit{agreements}. 
The cluster names were verified by two domain experts. 
We define each aspect below, accompanied by an illustrative example.

\begin{itemize}
    \item \textbf{Contact}: refers to the extent to which an employee can get in touch with the agency, for example, by phone or email. - \textit{It took a long time to receive a response from the contact person. I had to make multiple phone calls and send emails before getting a reply.}
    \item \textbf{Schedule}: is about scheduling, work hours, and days off and whether the agency is flexible in changing these. - \textit{I appreciate receiving my schedule well in advance because it allows me to adjust my plans accordingly.} 
    \item \textbf{Agreements}: relates to the arrangements made between the employee and the agency and whether these are upheld or not. - \textit{I am satisfied with the schedule agreement because it allows me to take my kids to school.}
    \item \textbf{Salary}: is about payment and any bonuses or extras such as travel expenses. Consider remarks about correct payment of salary, and the frequency of salary payment. - \textit{I am happy that I can choose my own frequency of payment.}
    \item \textbf{Personal attention}: is about the extent to which the agency pays personal attention to the employee. It can include receiving feedback and receiving personal guidance. - \textit{I feel valued as an employee, and my ideas and suggestions are listened to attentively.}
    \item \textbf{Communication}: refers to the way information is exchanged between the agency and the employee, and whether communication about important matters is timely. - \textit{I am satisfied with the way in which I am informed about the changes that are happening within the company.}
\end{itemize}

The \textit{contact} and \textit{communication} aspects may appear similar but are distinct: 
both revolve around interaction, but \textit{contact} pertains to the accessibility and responsiveness between employees and the agency, emphasizing ease of access and availability. 
\textit{Communication} is about information exchange, including factors like clarity, completeness, and timeliness. 

\begin{table*}[ht!]
\resizebox{\textwidth}{!}{%
\begin{tabular}{cllllllllllll}
\hline
Cluster     & \multicolumn{2}{c}{contact}                                                                                                                                                                      & \multicolumn{2}{c}{salary}                                                                                                                                                 & \multicolumn{2}{c}{schedule}                                                                                                                                & \multicolumn{2}{c}{personal attention}                                                                                                                                      & \multicolumn{2}{c}{communication}                                                                                                                                                     & \multicolumn{2}{c}{agreements}                                                                                                                                                                      \\ \hline
Top 5 terms & \begin{tabular}[c]{@{}l@{}}contact\\ persoon\\ opnemen\\ vraag\\ contactpersoon\end{tabular} & \begin{tabular}[c]{@{}l@{}}contact\\ person\\ to pick up\\ question\\ contact person\end{tabular} & \begin{tabular}[c]{@{}l@{}}krijgen\\ komen\\ willlen\\ vragen\\ jaar\end{tabular} & \begin{tabular}[c]{@{}l@{}}to receive\\ to come\\ to want\\ to ask\\ year\end{tabular} & \begin{tabular}[c]{@{}l@{}}week\\ dag\\ uur\\ maand\\ krijgen\end{tabular} & \begin{tabular}[c]{@{}l@{}}week\\ day\\ hour\\ month\\ to receive\end{tabular} & \begin{tabular}[c]{@{}l@{}}mens\\ contact\\ krijgen\\ nummer\\ maken\end{tabular} & \begin{tabular}[c]{@{}l@{}}human\\ contact\\ to receive\\ number\\ to make\end{tabular} & \begin{tabular}[c]{@{}l@{}}communicatie\\ verlopen\\ contact\\ komen\\ super\end{tabular} & \begin{tabular}[c]{@{}l@{}}communication\\ to go\\ contact\\ to come\\ super\end{tabular} & \begin{tabular}[c]{@{}l@{}}gesprek\\ horen\\ evaluatie\\ bellen\\ sollicitatie\end{tabular} & \begin{tabular}[c]{@{}l@{}}conversation\\ to hear\\ evaluation\\ to call\\ job interview\end{tabular} \\ \hline
\end{tabular}%
}
\caption{\small{Six identified aspects obtained through clustering responses, with the top five terms with highest TF-IDF scores.}}
\label{tab:toptermspercluster}
\end{table*}

\subsection{Annotation Study}\label{sec:annotationstudy}
To collect labelled data, we ran an annotation study with nine native Dutch-speaking trainees from Yacht, which is one of the staffing agencies within Randstad. 
We had each response in the set of 1,500 responses annotated by three annotators, i.e., each annotator annotated 500 responses. 

Annotators attended an in-person session to familiarize themselves with the task and guidelines. 
Detailed written guidelines, encompassing the task's objective, annotation procedure, answering options, and category definitions, were provided, along with examples for each category.
To address annotator bias, a preliminary sample of 20 responses was annotated and discussed before each annotator worked on their assigned batch. 
We have published an English translation of our annotation guidelines online.\footnote{\url{https://anonymous.4open.science/r/AnnotationGuidelinesABSA-BB08/}}

We employed the Prodigy annotation tool ~\cite{Prodigy:2018} with a custom recipe for annotation purposes. 
Annotators individually reviewed responses, selected relevant aspects, and a binary (positive or negative) sentiment for each aspect. 
They could also select 'no topics' if none of the six aspects were discussed. 
Responses were presented in random order to ensure unbiased judgment, and disagreements among annotators were resolved through majority voting, relying on a fourth annotator re-annotating in cases of no consensus. 

Our primary focus was to identify clear-cut positive and negative sentiments for actionable insights on employee satisfaction. 
To address conflicting sentiments toward the same aspect, an 'ignore' option was introduced. 
Sentences marked with 'ignore' were excluded, enabling the model to focus on identifiable sentiment patterns. 
In addition, as sentiment was modeled as a binary variable, for neutral sentiments or multiple sentiments toward a single aspect, annotators used the 'ignore' option to ensure data consistency \cite{hartmann2023more}.

\subsubsection{Inter-Annotator Agreement}
Reliability in assessing inter-annotator agreement (IAA) is crucial. In this study, we employed Fleiss' kappa to measure agreement among multiple annotators, an extension of Cohen's kappa for more than two annotators \cite{fleiss1971measuring}.

With an average kappa score of $0.537$, we achieved a moderate level of IAA, which reflects reliable annotations \cite{dumitrache2015achieving} considering the inherent language ambiguity and inter-annotator disagreement. 

The kappa statistic can be strict, especially in a multi-label setting, as it does not reward partial overlaps between annotations.
Upon examining disagreement cases, we observed that when two out of three annotators agreed on the exact annotation, $53.92\%$ of the responses exhibited a partial overlap between the majority-vote annotation and the third one. Additionally, 206 annotations (13\%) required re-annotation by a fourth annotator due to three annotators providing different answers.

A quantitative analysis investigated disagreement patterns among aspects. 
Out of 106 responses with disagreement, a notable pattern emerged regarding the 'communication' aspect. 
In these cases, two annotators selected 'no topics,' while one chose 'communication:NEG.' 
This suggests a lack of clear demarcation in defining 'communication,' particularly in negative discussions. 
A similar pattern was observed for 'personal attention,' possibly due to its over-representation in the dataset. 
Additionally, aspects with the highest agreement also generated substantial disagreement, likely because of their high frequency in the dataset. 
We found similar patterns for aspects with low occurrence.

In conclusion, the annotation study resulted in a dataset of 1,500 responses. 
Among these, 42 responses were categorized as 'ignore' due to conflicting or neutral sentiments and were excluded. 
After their removal, the final dataset comprised 1,458 responses, with 267 discussing aspects positively and 1,091 featuring negative discussions. 
See Table \ref{tab:distributionlabelssentiments} for a detailed distribution of aspect and sentiment labels.

\begin{table}[h]
\centering
\resizebox{1\columnwidth}{!}{%
\begin{tabular}{lccc}
\hline
\textbf{Label}              & \textbf{POS\_Count} & \textbf{NEG\_Count} & \textbf{total} \\ \hline
\textbf{agreements}         & 8                   & 67                  & 75             \\
\textbf{communication}      & 33                  & 212                 & 245            \\
\textbf{contact}            & 57                  & 155                 & 212            \\
\textbf{personal attention} & 141                 & 370                 & 511            \\
\textbf{schedule}           & 5                   & 134                 & 139            \\
\textbf{salary}             & 23                  & 153                 & 176            \\
\textbf{no topics}          & 0                   & 0                   & 376            \\ \hline
\textbf{total}              & 267                 & 1091                & 1734           \\ \hline
\end{tabular}%
}
\caption{Distribution of aspects and sentiments. POS\_Count indicates the number of positive occurrences in the data set; NEG\_Count indicates the negatives.}
\label{tab:distributionlabelssentiments}
\end{table}

\subsection{Data Augmentation}
Table \ref{tab:distributionlabelssentiments} reveals significant label imbalance in both aspects and sentiments, e.g., ‘personal attention’ is disproportionately represented, occurring nearly five times more than ‘agreements’. 
In addition, the majority (approximately $78.77\%$) of aspects have a negative sentiment. 

To address this imbalance and prevent bias in the machine learning model, data augmentation was implemented using NLPaug \cite{ma2019nlpaug}. 
This approach involves generating contextual word embeddings using a BERT model and replacing some of the tokens in a sentence \cite{sarhan2022uu}. 
For an example of an original sentence and its augmented version: 
\textit{"De lonen zouden wel een keer flink omhoog mogen"} (“the wages could well do with a substantial increase”) is rephrased into \textit{"De salarissen zullen tot twintig keer dik omhoog moeten."}
Here, some tokens from the initial sentence were replaced with contextually analogous tokens to yield the augmented sentence. 
However, this carries the risk of creating ungrammatical augmented sentences, as in the example.

Data augmentation used RobBERT embeddings to increase label combination variety until each distinct label combination occurred at least 30 times. 
RobBERT was chosen for its wider training data set, enabling broader coverage of Dutch texts. 
Because the objective was to extract multiple aspect-sentiments from open-ended survey responses, augmentation focused on responses with two or more aspects. 
With a $30\%$ set augmentation probability and a maximum of $50$ tokens, data augmentation involved replacing up to $50$ tokens in a response. 
The augmented responses were added to the training data set, as shown in Table \ref{tab:distributionaugmentedlabelssentiments}, which displays the distribution of labels and sentiments across both the augmented and non-augmented training data set.
Although Table \ref{tab:distributionaugmentedlabelssentiments} demonstrates the additional training samples improved balance, a slight imbalance remains. 

\begin{table}[ht]
\centering
\resizebox{1\columnwidth}{!}{%
\begin{tabular}{lcccc}
\hline
\multicolumn{1}{c}{\textbf{}} & \multicolumn{2}{c}{\textbf{POS\_Count}} & \multicolumn{2}{c}{\textbf{NEG\_Count}} \\ \hline
\textbf{Label}                & original           & augmented          & original           & augmented          \\ \hline
\textbf{agreements}           & 6                  & 200                & 43                 & 516                \\
\textbf{communication}        & 21                 & 516                & 140                & 745                \\
\textbf{contact}              & 28                 & 432                & 111                & 803                \\
\textbf{personal attention}   & 101                & 550                & 239                & 716                \\
\textbf{schedule}             & 3                  & 98                 & 85                 & 420                \\
\textbf{salary}               & 13                 & 380                & 102                & 539                \\ \hline
\textbf{total}                & 172                & 2176               & 720                & 3739               \\ \hline
\end{tabular}%
}
\caption{Aspect and sentiment distribution of the original versus the augmented training data set.}
\label{tab:distributionaugmentedlabelssentiments}
\end{table}

\subsection{Model implementation}
For aspect-based sentiment analysis of open-ended survey responses, we propose a two-tiered approach.
The first step employs multi-label classification to determine the correct aspects for each response. 
The second utilizes the aspects identified by the first system as features, along with the response, and assigns binary sentiment labels to each aspect within the response. 
The adoption of a two-tiered approach serves a dual purpose. Firstly, it allows the model to concentrate exclusively on aspect identification in its initial step. This deliberate isolation permits the model to specialize in autonomously recognizing aspects before undertaking sentiment classification. 
Moreover, the two-tiered framework is strategically designed to mitigate challenges associated with data sparsity in the dataset. Given that certain aspects may possess limited training examples, a 12-class multi-label classification experiment could potentially yield inadequate representations for specific aspects. This limitation may compromise the model's capacity to generalize beyond the training data and perform optimally across the entire spectrum of aspect-sentiment classes. 

\subsubsection{Baselines}
\label{sec:baseline}
For aspect and sentiment classification we employ support vector machines (SVM), multilayer perceptron (MLP), and two Dutch BERT models (BERTje and RobBERT) in a zero-shot classification setting as baselines. 
We selected BERTje and RobBERT for their success in similar tasks, and the advantage of pre-training on a larger corpus of Dutch texts~\cite{cammel2020automatically, van2022analyzing, de2022sentemo}. 

We apply hyperparameter tuning on a validation set, relying on a 70/15/15 train/test/validation split. 
For SVM, we found C=1000 and gamma=0.01 to be the optimal hyperparameters.
For MLP, we found ReLU activation, Adam solver, and a hidden layer size of (256, 128) as optimal hyperparameters.
For BERTje and RobBERT, we did parameter tuning on the training set, since no actual training was done. 
For aspect classification, each open-ended response was paired with all possible aspects, generating six inputs per response. 
Tokenisation was performed using the BERTje and RobBERT model tokenisers, following the guidelines provided by \citet{de2019bertje} and \citet{delobelle2020robbert}. 
The network produced a probability vector of length six, indicating the likelihood of each aspect's presence in the response. 
Predicted class probabilities were initially notably below 0.5, perhaps due to the models' lack of training on the target domain data, so we tuned classification thresholds through a grid search, resulting in thresholds of 0.45 for BERTje and 0.37 for RobBERT. Maintaining a 0.5 threshold would have led to numerous false negatives.

For sentiment classification, we followed the same approach for SVM and MLP. 
For SVM we applied a linear kernel to fit the binary nature of the task. 
The optimal C parameter was found at 10. 
For MLP, we found ReLU activation, Adam solver, and a hidden layer size of (128, 64) to be optimal parameters. 
The preprocessed data was passed through the network, and the model output was a two-element vector representing positive and negative sentiment classes, with the sentiment having the highest value assigned as the predicted sentiment.

\subsubsection{Aspect classification}
In the aspect classification task, we fine-tune BERTje and RobBERT in a few-shot setting. 
To maintain consistent input length, batches of 16 tokenized samples were generated.
The data set was randomly shuffled before training to mitigate order bias. 
The neural network consisted of a 12-layered BERT model with a dropout layer (dropout=.3) for regularization, and as output layer a linear layer with six dimensions representing six aspects. 
As both models converged around epoch 10 and to avoid overfitting, we stopped training at 10 epochs~\cite{yu2019interpreting}, using Adam optimizer with a learning rate of 0.005.
Binary cross-entropy loss was calculated separately for each class with sigmoid activation and network weights were updated based on the total loss. This sigmoid activation approach allows for independent and interpretable probability estimates for the presence of each aspect, facilitating a comprehensive multi-label classification strategy.

\subsubsection{Sentiment classification}
For sentiment classification, BERTje and RobBERT were used in a few-shot classification setting. 
Input responses were padded, tokenized, and shuffled using a batch size of 4. 
Categorical aspect features were encoded using an embedding layer, and their embeddings were concatenated with the BERT embeddings of the responses to generate distinct sentiment predictions for each aspect. 

The neural network for sentiment classification consisted of a 12-layer BERT model followed by a ReLU layer for learning complex patterns~\cite{goodfellow2016deep}, with a dropout layer (dropout=.3), followed by a 2 dimensional linear layer for the binary sentiment labels. 

The model's performance was evaluated using cross-entropy loss after each iteration. 
A training function trained the model for 10 epochs with a learning rate of 0.005 using the Adam optimizer. 
For RobBERT's an BERTje's training loss and accuracy, we found they stabilized after the fourth epoch. 
However, we extended training to avoid premature stopping and underfitting~\cite{yu2019interpreting}. 
The loss continued to decrease until epoch 10, indicating no overfitting through extended training.

\subsection{Evaluation Metrics}\label{evaluation_metrics}
To evaluate aspect and sentiment classification tasks, we use the macro F1 score, which balances precision and recall and treats each category equally, mitigating the impact of larger classes. We also examine precision and recall to detect potential overfitting and underfitting as suggested by \citet{sokolova2009systematic}.
The significance of the results is assessed using the Wilcoxon signed-rank test for aspect classification and McNemar's test for sentiment classification.

%% file: sections/4-results.tex
\section{Results}
\label{sec:results}
In this section, we present the results of the experiments.
We compare performance of BERTje$_{fewshot}$ and RobBERT$_{fewshot}$ to our traditional baselines ($SVM$ and $MLP$), and zero-shot BERT baselines (BERTje$_{zeroshot}$ and RobBERT$_{zeroshot}$). 
In addition, we apply data augmentation for both (BERTje$_{fewshotDA}$ and RobBERT$_{fewshotDA}$).

\subsection{Aspect classification}
First, we turn to Table~\ref{tab:results_zeroshotBERTje_RobBERT_aspects}, which shows the performance of BERTje$_{zeroshot}$ and RobBERT$_{zeroshot}$. 
We note that, with $0.8793$ recall and $0.1682$ precision for RobBERT$_{zeroshot}$, and $0.9129$ recall and $0.1568$ precision for BERTje$_{zeroshot}$, both approaches show overprediction, which persists across all aspect categories. 

The zero-shot models struggled to establish a reliable decision boundary for classifying aspects, indicating ineffective transfer of pre-training knowledge to novel data. 
The Dutch BERT models lacked domain-specific knowledge, aligning with prior findings for microblog texts~\cite{chen2021zero}.

\begin{table}[t]
\centering
\resizebox{\columnwidth}{!}{%
\begin{tabular}{lcccccc}
\hline
                               & \multicolumn{2}{c}{\textbf{precision}} & \multicolumn{2}{c}{\textbf{recall}} & \multicolumn{2}{c}{\textbf{f1-score}} \\ \hline
                               & BERTje            & RobBERT            & BERTje           & RobBERT          & BERTje            & RobBERT           \\ \hline
\textbf{agreements}             & 0.0488            & 0.0784             & 1                & 0.9153           & 0.0928            & 0.1381            \\
\textbf{communication}          & 0.1648            & 0.1623             & 0.9875           & 0.8441           & 0.2883            & 0.2662            \\
\textbf{contact}               & 0.1478            & 0.1728             & 0.8394           & 0.8871           & 0.2575            & 0.2818            \\
\textbf{personal attention} & 0.3452            & 0.03591            & 0.892            & 0.8974           & 0.5084            & 0.513             \\
\textbf{schedule}  & 0.0972            & 0.1072             & 0.8871           & 0.9167           & 0.1702            & 0.1983            \\
\textbf{salary}               & 0.1283            & 0.1211             & 0.8954           & 0.8218           & 0.2148            & 0.2148            \\ \hline
\textbf{macro avg}             & 0.1568            & 0.1682             & 0.9129           & 0.8793           & 0.2571            & 0.2671            \\ \hline
\end{tabular}%
}
\caption{zero-shot aspect classification scores.}
\label{tab:results_zeroshotBERTje_RobBERT_aspects}
\end{table}

Next, we compare the F1 scores of few-shot methods to all others, in Table~\ref{tab:results_aspects}. 
We see how BERTje$_{fewshot}$ and RobBERT$_{fewshot}$ at $0.5219$ and $0.5449$ respectively, significantly outperform all baselines ($p<0.0001$). 
Between them, RobBERT$_{fewshot}$ significantly outperforms BERTje$_{fewshot}$ ($p<0.0001$).

Our BoW baselines outperform the transformer-based zero-shot baselines, which suggest that the individual words captured by BoW models have a strong correlation with the aspects, and the contextual knowledge from BERT models may not offer sufficient information for distinguishing between aspects in open-ended survey responses. 

We applied data augmentation to address label imbalance. 
BERTje$_{fewshotDA}$ achieves a lower F1 score ($0.4982$) than BERTje$_{fewshot}$ ($0.5219$), which suggests potential overfitting, or that the model gained limited novel information from augmented examples. 
However, RobBERT$_{fewshotDA}$ outperforms RobBERT$_{fewshot}$ with a significant increase in F1 score from $0.5449$ to $0.6074$ ($p=0.017$). 
The performance improvement was particularly prominent in the ‘agreements’ category, which was underrepresented before augmentation. 
Nevertheless, data augmentation leads to decreased performance in some aspects. 

\begin{table*}
\centering
\resizebox{\textwidth}{!}{%
\begin{tabular}{lcccccccc}
\hline
\multicolumn{1}{c}{}        & \textbf{SVM} & \textbf{MLP} & \textbf{\begin{tabular}[c]{@{}c@{}}BERTje\\ $_{zeroshot}$\end{tabular}} & \textbf{\begin{tabular}[c]{@{}c@{}}RobBERT\\ $_{zeroshot}$\end{tabular}} & \textbf{\begin{tabular}[c]{@{}c@{}}BERTje\\ $_{fewshot}$\end{tabular}} & \textbf{\begin{tabular}[c]{@{}c@{}}BERTje\\ $_{fewshotDA}$\end{tabular}} & \textbf{\begin{tabular}[c]{@{}c@{}}RobBERT\\ $_{fewshot}$\end{tabular}} & \textbf{\begin{tabular}[c]{@{}c@{}}RobBERT\\ $_{fewshotDA}$\end{tabular}} \\ \hline
\textbf{agreements}         & 0.2963       & 0.2222       & 0.0928                                                                  & 0.1356                                                                  & 0.2849                                                                 & 0.2871                                                                   & 0.1553                                                                  & \textbf{0.4765 }                                                                   \\
\textbf{communication}      & 0.2857       & 0.1924       & 0.2883                                                                  & 0.2716                                                                  & 0.437                                                                  & 0.4298                                                                   & 0.3938                                                                  & \textbf{0.4691}                                                                    \\
\textbf{contact}            & 0.3619       & 0.4143       & 0.2575                                                                  & 0.2841                                                                  & 0.56                                                                   & \textbf{0.5984}                                                                   & 0.5758                                                                  & 0.5546                                                                    \\
\textbf{personal attention} & 0.6025       & 0.5871       & 0.5084                                                                  & 0.5148                                                                  & 0.6324                                                                 & 0.7064                                                                   & \textbf{0.7593}                                                                  & 0.6587                                                                    \\
\textbf{schedule}           & 0.4          & 0.3238       & 0.1702                                                                  & 0.2019                                                                  & 0.4892                                                                 & 0.3581                                                                   & 0.6129                                                                  & \textbf{0.7489}                                                                    \\
\textbf{salary}             & 0.5227       & 0.4498       & 0.2148                                                                  & 0.2142                                                                  & 0.7329                                                                 & 0.6157                                                                   & \textbf{0.7581}                                                                  & 0.6487                                                                    \\ \hline
\textbf{macro average}      & 0.4115       & 0.3633       & 0.2571                                                                  & 0.2671                                                                  & 0.5219                                                                 & 0.4982                                                                   & 0.5449                                                                  & \textbf{0.6074}                                                                    \\ \hline
\end{tabular}%
}
\caption{Aspect classification performance in terms of F1 score for; best performing methods are boldfaced.}
\label{tab:results_aspects}
\end{table*}

Both zero-shot models suffer from overprediction, evidenced by high recall and low precision. 
This could be caused by a lack of knowledge from the target domain. 
This finding is supported by the outcomes of the few-shot classification experiment, where the significantly improved performance shows the model's improved capability to differentiate between the different aspect-classes, resulting in a higher macro F1 score. This illustrates the importance and benefits of fine-tuning, even when only a small amount of labelled target domain data is available.\\

\subsection{Sentiment classification}
Turning to sentiment classification results in Table \ref{tab:results_sentiments}, we see how BERTje$_{fewshot}$ with an F1 score of $0.8736$ does not significantly outperform RobBERT$_{fewshot}$ at $0.8871$ ($p=0.292$). 
Both significantly outperform all baselines (all $p<0.0001$). 
Again, BoW models outperform zero-shot models that seem to struggle to transfer contextual knowledge to the novel domain. 

\begin{table*}[ht]
\centering
\resizebox{\textwidth}{!}{%
\begin{tabular}{lcccccccc}
\hline
\multicolumn{1}{c}{}   & \textbf{SVM} & \textbf{MLP} & \textbf{\begin{tabular}[c]{@{}c@{}}BERTje\\ $_{zeroshot}$\end{tabular}} & \textbf{\begin{tabular}[c]{@{}c@{}}RobBERT\\ $_{zershot}$\end{tabular}} & \textbf{\begin{tabular}[c]{@{}c@{}}BERTje\\ $_{fewshot}$\end{tabular}} & \textbf{\begin{tabular}[c]{@{}c@{}}BERTje\\ $_{fewshotDA}$\end{tabular}} & \textbf{\begin{tabular}[c]{@{}c@{}}RobBERT\\ $_{fewshot}$\end{tabular}} & \textbf{\begin{tabular}[c]{@{}c@{}}RobBERT\\ $_{fewshotDA}$\end{tabular}} \\ \hline
\textbf{negative}      & 0.9124       & 0.9024       & 0.2619                                                                  & 0.2938                                                                  & 0.9472                                                                 & 0.9423                                                                   & \textbf{0.9576}                                                                  & 0.9482                                                                    \\
\textbf{positive}      & 0.6228       & 0.5135       & 0.3348                                                                  & 0.3171                                                                  & 0.8                                                                    & 0.7912                                                                   & \textbf{0.8314}                                                                  & 0.8186                                                                    \\ \hline
\textbf{macro average} & 0.7676       & 0.708        & 0.2983                                                                  & 0.3061                                                                  & 0.8736                                                                 & 0.8651                                                                   & \textbf{0.8871}                                                                  & 0.8846                                                                    \\ \hline
\end{tabular}%
}
\caption{Sentiment classification performance in terms of F1 score; best performing methods are boldfaced.}
\label{tab:results_sentiments}
\end{table*}

There seems to be no beneficial impact of data augmentation on sentiment classification, with neither BERTje$_{fewshotDA}$ nor RobBERT$_{fewshotDA}$ being able to outperform BERTje$_{fewshot}$ ($0.8736$ vs. $0.8651$) and RobBERT$_{fewshot}$ ($0.8871$ vs. $0.8846$).

%% file: sections/5-discussion.tex
\section{Discussion}
\label{sec:discussion}
This study explores ABSA in Dutch employee satisfaction surveys, using Dutch BERT-based machine learning models. 
Our findings are in line with findings in prior research \cite{chang2020taming, karl2022transformers} that highlight BERT's effectiveness for ABSA in English. 
Consistent with \citet{karl2022transformers}, RobBERT outperforms BERTje, indicating the superiority of larger transformer models for sentiment classification, also observed by \citet{de2021emotional}.

Additionally, our study underscores the success of few-shot classification in addressing limited labelled data, consistent with \citet{hu2021multi} for aspect classification and \citet{dogra2021analyzing} for sentiment classification in English. 
However, our findings contradict successful application of BERT models for zero-shot classification by \citet{yin2019benchmarking}. 
This discrepancy in performance can be attributed to our domain-specific data, in contrast to the diverse dataset used by \citet{yin2019benchmarking}. 

Furthermore, this study identified significant label imbalance in aspects and their associated sentiments, as detailed in Section \ref{sec:results}. To address this, we explored data augmentation using NLPaug \cite{ma2019nlpaug} following \citet{sarhan2022uu}. However, this technique improved the macro F1 score for RobBERT in aspect classification only.

\subsection{Limitations}
Our dataset, comprising 1,458 responses, is relatively small which may affect its reliability. 
Despite time constraints, three annotators assessed each response for inter-annotator agreement. 
However, using larger and more diverse datasets can improve findings in future studies.

Using $k$-means clustering for aspect identification presents inherent limitations for internal and external validity. 
A substantial number of responses (376) didn't align with identified aspects, raising concerns about their reliability and comprehensiveness. 
Clustering does not support responses' potential membership of multiple clusters. 
Fuzzy clustering, as suggested by \citet{zhao2017fuzzy}, allows responses to belong to multiple clusters with varying membership degrees, offering a more comprehensive solution, at the cost of hindering clear boundaries and optimal cluster determination through the elbow method.
Finally, using a supervised classification approach on an internal dataset provides accurate clustering results but reduces findings' transferability to emerging aspects, impacting external validity. 
Unsupervised clustering, as demonstrated by \citet{cammel2020automatically}, may provide flexibility and adaptability to evolving contexts and domains.

To ensure high performance on future open-ended responses, further fine-tuning or retraining of the model with recent and relevant data is necessary. 

%% file: sections/6-conclusion.tex
\section{Conclusion} \label{sec:conclusion}
Analyzing workforce opinions and preferences through aspect-based sentiment analysis has various HR applications. In this paper, we demonstrate the effectiveness of Dutch BERT models, BERTje and RobBERT, in a few-shot ABSA experiment using Dutch open-ended responses from employee satisfaction surveys.
We address three sub-questions to gain insights into the models' performance and potential enhancements for aspect-based sentiment analysis (ABSA).

Regarding the first two sub-questions (RQ1.1 and RQ1.2), few-shot transformer models outperform baseline BoW models, significantly improving aspect-sentiment classification measured by macro F1 score. This emphasizes the importance of labeled data for fine-tuning BERT models, enhancing performance compared to relying solely on pre-trained knowledge in zero-shot scenarios. It also highlights BoW models' superior performance in leveraging individual words compared to zero-shot models struggling with domain transfer.

Regarding the third sub-research question (RQ1.3), this study demonstrates how data augmentation can enhance RobBERT's aspect classification performance. However, data augmentation does not improve aspect or sentiment classification for BERTje or sentiment classification for RobBERT. These findings suggest that the effectiveness of data augmentation varies across models and tasks.

In summary, regarding the main research question (RQ1) on BERT-based models' effectiveness in extracting aspect-sentiments, our findings demonstrate their superiority over traditional bag-of-words models and zero-shot classification approaches.
To enhance model robustness, future studies should acquire larger and more diverse ABSA datasets, exposing models to varied open-ended survey responses for improved generalization to novel data. 
Considering the challenges posed by the limitations of traditional clustering methods, future studies could explore the incorporation of other clustering methods such as fuzzy clustering ~\cite{zhao2017fuzzy} or semi-supervised (neural) topic modeling approaches ~\cite{chiu-etal-2022-joint, xu-etal-2023-vontss}. 
Moreover, future studies should investigate the disparity in data augmentation success between BERTje and RobBERT. Specifically, exploring whether generating augmented sentences using BERTje embeddings improves BERTje model performance, similar to the favorable outcome observed for RobBERT in this study. Understanding such disparities would contribute to a deeper understanding of the relationship between specific models, their embeddings, and the efficiency of data augmentation techniques in ABSA's broader context.

%% file: acl_latex.bbl
\begin{thebibliography}{40}
\expandafter\ifx\csname natexlab\endcsname\relax\def\natexlab#1{#1}\fi

\bibitem[{Bogers et~al.(2022)Bogers, Graus, Kaya, Guti\'{e}rrez, Mesbah, and
  Johnson}]{10.1145/3523227.3547414}
Toine Bogers, David Graus, Mesut Kaya, Francisco Guti\'{e}rrez, Sepideh Mesbah,
  and Chris Johnson. 2022.
\newblock \href {https://doi.org/10.1145/3523227.3547414} {Second workshop on
  recommender systems for human resources (recsys in hr 2022)}.
\newblock In \emph{Proceedings of the 16th ACM Conference on Recommender
  Systems}, RecSys '22, page 671–674, New York, NY, USA. Association for
  Computing Machinery.

\bibitem[{Boguraev et~al.(1999)Boguraev, Bellamy, and
  Kennedy}]{boguraev1999dynamic}
Branimir Boguraev, Rachel Bellamy, and Christopher Kennedy. 1999.
\newblock Dynamic presentation of phrasally-based document abstractions.
\newblock In \emph{Proceedings of the 32nd Annual Hawaii International
  Conference on Systems Sciences. 1999. HICSS-32. Abstracts and CD-ROM of Full
  Papers}, pages 10--pp. IEEE.

\bibitem[{Cammel et~al.(2020)Cammel, De~Vos, van Soest, Hettne, Boer,
  Steyerberg, and Boosman}]{cammel2020automatically}
Simone~A Cammel, Marit~S De~Vos, Daphne van Soest, Kristina~M Hettne, Fred
  Boer, Ewout~W Steyerberg, and Hileen Boosman. 2020.
\newblock How to automatically turn patient experience free-text responses into
  actionable insights: a natural language programming (nlp) approach.
\newblock \emph{BMC medical informatics and decision making}, 20(1):1--10.

\bibitem[{Chang et~al.(2020)Chang, Yu, Zhong, Yang, and
  Dhillon}]{chang2020taming}
Wei-Cheng Chang, Hsiang-Fu Yu, Kai Zhong, Yiming Yang, and Inderjit~S Dhillon.
  2020.
\newblock Taming pretrained transformers for extreme multi-label text
  classification.
\newblock In \emph{Proceedings of the 26th ACM SIGKDD international conference
  on knowledge discovery \& data mining}, pages 3163--3171.

\bibitem[{Chen et~al.(2021)Chen, Wang, Huang, and Coenen}]{chen2021zero}
Qi~Chen, Wei Wang, Kaizhu Huang, and Frans Coenen. 2021.
\newblock Zero-shot text classification via knowledge graph embedding for
  social media data.
\newblock \emph{IEEE Internet of Things Journal}, 9(12):9205--9213.

\bibitem[{Chiu et~al.(2022)Chiu, Mittal, Tumma, Sharma, and
  Doshi-Velez}]{chiu-etal-2022-joint}
Jeffrey Chiu, Rajat Mittal, Neehal Tumma, Abhishek Sharma, and Finale
  Doshi-Velez. 2022.
\newblock \href {https://doi.org/10.18653/v1/2022.spnlp-1.5} {A joint learning
  approach for semi-supervised neural topic modeling}.
\newblock In \emph{Proceedings of the Sixth Workshop on Structured Prediction
  for NLP}, pages 40--51, Dublin, Ireland. Association for Computational
  Linguistics.

\bibitem[{Dadgar et~al.(2016)Dadgar, Araghi, and Farahani}]{dadgar2016novel}
Seyyed Mohammad~Hossein Dadgar, Mohammad~Shirzad Araghi, and Morteza~Mastery
  Farahani. 2016.
\newblock A novel text mining approach based on tf-idf and support vector
  machine for news classification.
\newblock In \emph{2016 IEEE International Conference on Engineering and
  Technology (ICETECH)}, pages 112--116. IEEE.

\bibitem[{De~Bruyne et~al.(2021)De~Bruyne, De~Clercq, and
  Hoste}]{de2021emotional}
Luna De~Bruyne, Orph{\'e}e De~Clercq, and V{\'e}ronique Hoste. 2021.
\newblock Emotional robbert and insensitive bertje: combining transformers and
  affect lexica for dutch emotion detection.
\newblock In \emph{Workshop on Computational Approaches to Subjectivity and
  Sentiment Analysis (WASSA), held in conjunction with EACL 2021}, pages
  257--263. Association for Computational Linguistics.

\bibitem[{De~Clercq and Hoste(2016)}]{de2016rude}
Orph{\'e}e De~Clercq and V{\'e}ronique Hoste. 2016.
\newblock Rude waiter but mouthwatering pastries! an exploratory study into
  dutch aspect-based sentiment analysis.
\newblock In \emph{Proceedings of the Tenth International Conference on
  Language Resources and Evaluation (LREC'16)}, pages 2910--2917.

\bibitem[{De~Geyndt et~al.(2022)De~Geyndt, De~Clercq, Van~Hee, Lefever, Singh,
  Parent, and Hoste}]{de2022sentemo}
Ellen De~Geyndt, Orph{\'e}e De~Clercq, Cynthia Van~Hee, Els Lefever, Pranaydeep
  Singh, Olivier Parent, and Veronique Hoste. 2022.
\newblock Sentemo: A multilingual adaptive platform for aspect-based sentiment
  and emotion analysis.
\newblock In \emph{12th Workshop on Computational Approaches to Subjectivity,
  Sentiment \& Social Media Analysis, collocated with ACL 2022}, pages 51--61.
  Association for Computational Linguistics.

\bibitem[{de~Vries et~al.(2019)de~Vries, van Cranenburgh, Bisazza, Caselli,
  Noord, and Nissim}]{devries2019bertje}
Wietse de~Vries, Andreas van Cranenburgh, Arianna Bisazza, Tommaso Caselli,
  Gertjan~van Noord, and Malvina Nissim. 2019.
\newblock \href {http://arxiv.org/abs/1912.09582} {{BERTje}: {A} {Dutch} {BERT}
  {Model}}.
\newblock arXiv:1912.09582.

\bibitem[{De~Vries et~al.(2019)De~Vries, van Cranenburgh, Bisazza, Caselli, van
  Noord, and Nissim}]{de2019bertje}
Wietse De~Vries, Andreas van Cranenburgh, Arianna Bisazza, Tommaso Caselli,
  Gertjan van Noord, and Malvina Nissim. 2019.
\newblock Bertje: A dutch bert model.
\newblock \emph{arXiv preprint arXiv:1912.09582}.

\bibitem[{Delobelle et~al.(2020)Delobelle, Winters, and
  Berendt}]{delobelle2020robbert}
Pieter Delobelle, Thomas Winters, and Bettina Berendt. 2020.
\newblock Robbert: a dutch roberta-based language model.
\newblock \emph{arXiv preprint arXiv:2001.06286}.

\bibitem[{Dogra et~al.(2021)Dogra, Singh, Verma, Kavita, Jhanjhi, and
  Talib}]{dogra2021analyzing}
Varun Dogra, Aman Singh, Sahil Verma, Kavita, NZ~Jhanjhi, and MN~Talib. 2021.
\newblock Analyzing distilbert for sentiment classification of banking
  financial news.
\newblock In \emph{Intelligent Computing and Innovation on Data Science:
  Proceedings of ICTIDS 2021}, pages 501--510. Springer.

\bibitem[{Dumitrache et~al.(2015)Dumitrache, Aroyo, and
  Welty}]{dumitrache2015achieving}
Anca Dumitrache, Lora Aroyo, and Chris Welty. 2015.
\newblock Achieving expert-level annotation quality with crowdtruth.
\newblock In \emph{Proc. of BDM2I Workshop, ISWC}.

\bibitem[{Fleiss(1971)}]{fleiss1971measuring}
Joseph~L Fleiss. 1971.
\newblock Measuring nominal scale agreement among many raters.
\newblock \emph{Psychological bulletin}, 76(5):378.

\bibitem[{Goodfellow et~al.(2016)Goodfellow, Bengio, and
  Courville}]{goodfellow2016deep}
Ian Goodfellow, Yoshua Bengio, and Aaron Courville. 2016.
\newblock \emph{Deep learning}.
\newblock MIT press.

\bibitem[{Hartmann et~al.(2023)Hartmann, Heitmann, Siebert, and
  Schamp}]{hartmann2023more}
Jochen Hartmann, Mark Heitmann, Christian Siebert, and Christina Schamp. 2023.
\newblock More than a feeling: Accuracy and application of sentiment analysis.
\newblock \emph{International Journal of Research in Marketing}, 40(1):75--87.

\bibitem[{Hoang et~al.(2019)Hoang, Bihorac, and Rouces}]{hoang2019aspect}
Mickel Hoang, Oskar~Alija Bihorac, and Jacobo Rouces. 2019.
\newblock Aspect-based sentiment analysis using bert.
\newblock In \emph{Proceedings of the 22nd nordic conference on computational
  linguistics}, pages 187--196.

\bibitem[{Hosseini-Asl et~al.(2022)Hosseini-Asl, Liu, and
  Xiong}]{hosseini2022generative}
Ehsan Hosseini-Asl, Wenhao Liu, and Caiming Xiong. 2022.
\newblock A generative language model for few-shot aspect-based sentiment
  analysis.
\newblock \emph{arXiv preprint arXiv:2204.05356}.

\bibitem[{Hu et~al.(2021)Hu, Zhao, Guo, Xue, Gao, Gao, Cheng, and
  Su}]{hu2021multi}
Mengting Hu, Shiwan Zhao, Honglei Guo, Chao Xue, Hang Gao, Tiegang Gao, Renhong
  Cheng, and Zhong Su. 2021.
\newblock Multi-label few-shot learning for aspect category detection.
\newblock \emph{arXiv preprint arXiv:2105.14174}.

\bibitem[{Jim{\'e}nez-Zafra et~al.(2017)Jim{\'e}nez-Zafra,
  Mart{\'\i}n-Valdivia, Maks, and Izquierdo}]{jimenez2017analysis}
Salud~Mar{\'\i}a Jim{\'e}nez-Zafra, M~Teresa Mart{\'\i}n-Valdivia, Isa Maks,
  and Rub{\'e}n Izquierdo. 2017.
\newblock Analysis of patient satisfaction in dutch and spanish online reviews.
\newblock \emph{Procesamiento del Lenguaje Natural}, 58:101--108.

\bibitem[{Karl and Scherp(2022)}]{karl2022transformers}
Fabian Karl and Ansgar Scherp. 2022.
\newblock Transformers are short text classifiers: A study of inductive short
  text classifiers on benchmarks and real-world datasets.
\newblock \emph{arXiv preprint arXiv:2211.16878}.

\bibitem[{Kumar(2019)}]{Kumar2019Sentiment}
H.~Kumar. 2019.
\newblock Sentiment analysis on imdb movie reviews using hybrid feature
  extraction method.
\newblock \emph{Int. J. Interact. Multim. Artif. Intell.}

\bibitem[{Liao et~al.(2021)Liao, Zeng, Yin, and Wei}]{liao2021improved}
Wenxiong Liao, Bi~Zeng, Xiuwen Yin, and Pengfei Wei. 2021.
\newblock An improved aspect-category sentiment analysis model for text
  sentiment analysis based on roberta.
\newblock \emph{Applied Intelligence}, 51:3522--3533.

\bibitem[{Lin and He(2009)}]{lin2009joint}
Chenghua Lin and Yulan He. 2009.
\newblock Joint sentiment/topic model for sentiment analysis.
\newblock In \emph{Proceedings of the 18th ACM conference on Information and
  knowledge management}, pages 375--384.

\bibitem[{Ma(2019)}]{ma2019nlpaug}
Edward Ma. 2019.
\newblock Nlp augmentation.
\newblock https://github.com/makcedward/nlpaug.

\bibitem[{Montani and Honnibal(2018)}]{Prodigy:2018}
Ines Montani and Matthew Honnibal. 2018.
\newblock \href {http://arxiv.org/abs/to appear} {Prodigy: A new annotation
  tool for radically efficient machine teaching}.
\newblock \emph{Artificial Intelligence}, to appear.

\bibitem[{Nazir et~al.(2020)Nazir, Rao, Wu, and Sun}]{nazir2020issues}
Ambreen Nazir, Yuan Rao, Lianwei Wu, and Ling Sun. 2020.
\newblock Issues and challenges of aspect-based sentiment analysis: A
  comprehensive survey.
\newblock \emph{IEEE Transactions on Affective Computing}, 13(2):845--863.

\bibitem[{Pontiki et~al.(2016)Pontiki, Galanis, Papageorgiou, Androutsopoulos,
  Manandhar, AL-Smadi, Al-Ayyoub, Zhao, Qin, De~Clercq
  et~al.}]{pontiki2016semeval}
Maria Pontiki, Dimitris Galanis, Haris Papageorgiou, Ion Androutsopoulos,
  Suresh Manandhar, Mohammed AL-Smadi, Mahmoud Al-Ayyoub, Yanyan Zhao, Bing
  Qin, Orph{\'e}e De~Clercq, et~al. 2016.
\newblock Semeval-2016 task 5: Aspect based sentiment analysis.
\newblock In \emph{ProWorkshop on Semantic Evaluation (SemEval-2016)}, pages
  19--30. Association for Computational Linguistics.

\bibitem[{Pontiki et~al.(2014)Pontiki, Papageorgiou, Galanis, Androutsopoulos,
  Pavlopoulos, and Manandhar}]{pontiki2014semeval}
Maria Pontiki, Haris Papageorgiou, Dimitrios Galanis, Ion Androutsopoulos, John
  Pavlopoulos, and Suresh Manandhar. 2014.
\newblock Semeval-2014 task 4: Aspect based sentiment analysis.
\newblock \emph{SemEval 2014}, page~27.

\bibitem[{Sarhan et~al.(2022)Sarhan, Mosteiro, and Spruit}]{sarhan2022uu}
Injy Sarhan, Pablo Mosteiro, and Marco Spruit. 2022.
\newblock Uu-tax at semeval-2022 task 3: Improving the generalizability of
  language models for taxonomy classification through data augmentation.
\newblock \emph{arXiv preprint arXiv:2210.03378}.

\bibitem[{Sokolova and Lapalme(2009)}]{sokolova2009systematic}
Marina Sokolova and Guy Lapalme. 2009.
\newblock A systematic analysis of performance measures for classification
  tasks.
\newblock \emph{Information processing \& management}, 45(4):427--437.

\bibitem[{Tesfagergish et~al.(2022)Tesfagergish,
  Kapo{\v{c}}i{\=u}t{\.e}-Dzikien{\.e}, and
  Dama{\v{s}}evi{\v{c}}ius}]{tesfagergish2022zero}
Senait~Gebremichael Tesfagergish, Jurgita Kapo{\v{c}}i{\=u}t{\.e}-Dzikien{\.e},
  and Robertas Dama{\v{s}}evi{\v{c}}ius. 2022.
\newblock Zero-shot emotion detection for semi-supervised sentiment analysis
  using sentence transformers and ensemble learning.
\newblock \emph{Applied Sciences}, 12(17):8662.

\bibitem[{van Buchem et~al.(2022)van Buchem, Neve, Kant, Steyerberg, Boosman,
  and Hensen}]{van2022analyzing}
Marieke~M van Buchem, Olaf~M Neve, Ilse~MJ Kant, Ewout~W Steyerberg, Hileen
  Boosman, and Erik~F Hensen. 2022.
\newblock Analyzing patient experiences using natural language processing:
  development and validation of the artificial intelligence patient reported
  experience measure (ai-prem).
\newblock \emph{BMC Medical Informatics and Decision Making}, 22(1):1--11.

\bibitem[{Wu(2020)}]{Wu2020Identifying}
Jheng-Long Wu. 2020.
\newblock Identifying emotion labels from psychiatric social texts using a
  bi-directional lstm-cnn model.
\newblock \emph{IEEE Access}.

\bibitem[{Xu et~al.(2023)Xu, Jiang, Sengamedu Hanumantha~Rao, Iannacci, and
  Zhao}]{xu-etal-2023-vontss}
Weijie Xu, Xiaoyu Jiang, Srinivasan Sengamedu Hanumantha~Rao, Francis Iannacci,
  and Jinjin Zhao. 2023.
\newblock \href {https://doi.org/10.18653/v1/2023.findings-acl.271} {v{ONTSS}:
  v{MF} based semi-supervised neural topic modeling with optimal transport}.
\newblock In \emph{Findings of the Association for Computational Linguistics:
  ACL 2023}, pages 4433--4457, Toronto, Canada. Association for Computational
  Linguistics.

\bibitem[{Yin et~al.(2019)Yin, Hay, and Roth}]{yin2019benchmarking}
Wenpeng Yin, Jamaal Hay, and Dan Roth. 2019.
\newblock Benchmarking zero-shot text classification: Datasets, evaluation and
  entailment approach.
\newblock \emph{arXiv preprint arXiv:1909.00161}.

\bibitem[{Yu et~al.(2019)Yu, Qin, Liu, Zhao, Wang, and
  Chen}]{yu2019interpreting}
Fuxun Yu, Zhuwei Qin, Chenchen Liu, Liang Zhao, Yanzhi Wang, and Xiang Chen.
  2019.
\newblock Interpreting and evaluating neural network robustness.
\newblock \emph{arXiv preprint arXiv:1905.04270}.

\bibitem[{Zhao and Mao(2017)}]{zhao2017fuzzy}
Rui Zhao and Kezhi Mao. 2017.
\newblock Fuzzy bag-of-words model for document representation.
\newblock \emph{IEEE transactions on fuzzy systems}, 26(2):794--804.

\end{thebibliography}
